\documentclass[conference]{IEEEtran}
\IEEEoverridecommandlockouts

\usepackage{cite}
\usepackage{amsmath,amssymb,amsfonts}
\usepackage{algorithmic}
\usepackage{graphicx}
\usepackage{textcomp}
\usepackage{xcolor}
\usepackage{siunitx}
\def\BibTeX{{\rm B\kern-.05em{\sc i\kern-.025em b}\kern-.08em
    T\kern-.1667em\lower.7ex\hbox{E}\kern-.125emX}}

\usepackage{color}
\usepackage{xcolor}
\newcommand{\minisection}[1]{\vspace{0.05in} \noindent {\bf #1} \ }

\usepackage{todonotes}
\usepackage[numbers]{natbib}

\begin{document}

\title{Augmenting Game AI with Deep Reinforcement Learning}

\author{\IEEEauthorblockN{Alessandro Sestini\textsuperscript{*}, Joakim Bergdahl, Amir Baghi, Jean-Philippe Barrette-LaPierre, Florian Fuchs, Linus Gisslén\textsuperscript{*}}
\IEEEauthorblockN{\textit{Electronic Arts (EA), Stockholm, Sweden}\\
 \{asestini,jbergdahl,abaghi,jbarrettelapierre,ffuchs,lgisslen\}@ea.com \\
 \textsuperscript{*}Corresponding authors
}}

\IEEEoverridecommandlockouts
\IEEEpubid{\makebox[\columnwidth]{* 979-8-3315-9476-3/26/\$31.00 \copyright2026 European Union\hfill} 
\hspace{\columnsep}\makebox[\columnwidth]{ }}

\maketitle

\IEEEpubidadjcol

\begin{abstract}
Immersion in video games depends not only on graphics, audio, and game mechanics, but also on the quality of in-game characters. Producing believable characters, or game AI, remains a significant challenge as behavioral complexity is hard to capture with hand-coded systems. Game AI is a source of immersion and engagement; however, the limitations stemming from the challenges of creating game AI often lead to frustration and the breaking of the illusion of realism within the game.
The introduction of machine learning models opens the door to creating more believable, authentic, and relatable characters in games. The promise is that they either learn from interacting with the game, or from player data, to develop true human-like behavior.
In this paper, we envision more applications of reinforcement learning for game AI in the future. For this to materialize, current research limitations are prohibitive to broad deployment across game genres. Therefore, we propose a framework for training reinforcement learning models with a set of requirements in mind that are suited towards game AI and game development. We present examples of games with reinforcement learning-augmented game AI and describe the practicalities of deploying player-facing machine learning agents in modern games. Furthermore, we identify bottlenecks and hard problems in these areas, which we believe offer promising research directions to accelerate the adoption of machine learning in game AI for the video game industry.
\end{abstract}


\section{Introduction}
\label{sec:introduction}
Machine Learning (ML), and in particular Reinforcement Learning (RL), has shown impressive results in playing a variety of games at human and superhuman levels, with notable examples such as Alphastar~\cite{vinyals2019grandmaster}, GT Sophy~\cite{wurman2022outracing}, and others~\cite{berner2019dota, bairamian2023minimax, wei2022honor}. More recently, the ability to play a video game from scratch with no human interactions has proven useful in video game production settings, particularly for gameplay testing~\cite{sestini2023towards, sestini2022automated}. For instance, RL-based agents have been employed in production to test releases of the \emph{Battlefield} series~\cite{gillberg2023technical}, and an increasing number of gaming companies are applying ML-based techniques specifically for testing purposes~\cite{modl, nunu}.

\begin{figure}
    \centering
    \begin{tabular}{c}
         \includegraphics[width=0.35\textwidth]{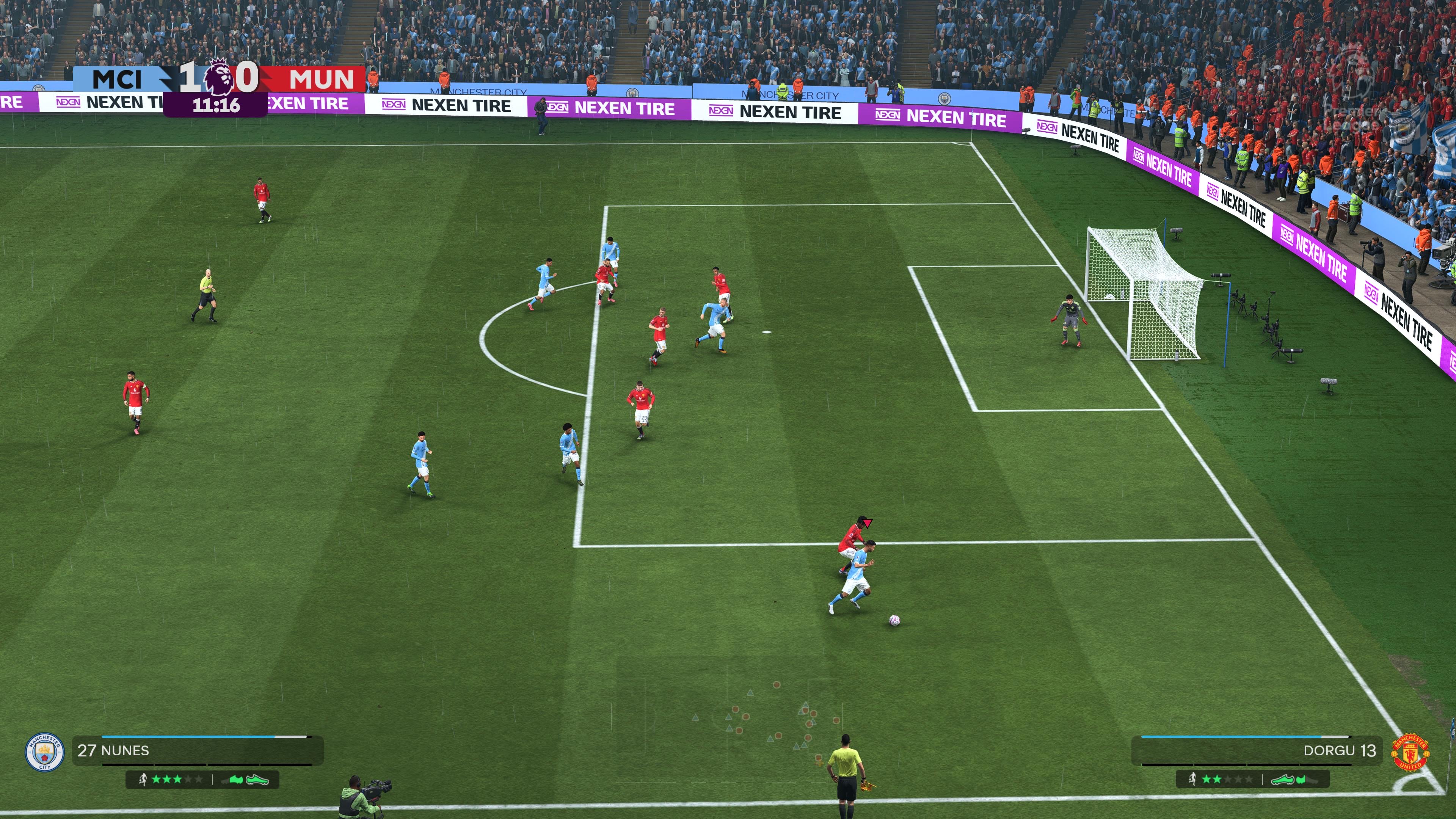} \\
         \includegraphics[width=0.35\textwidth]{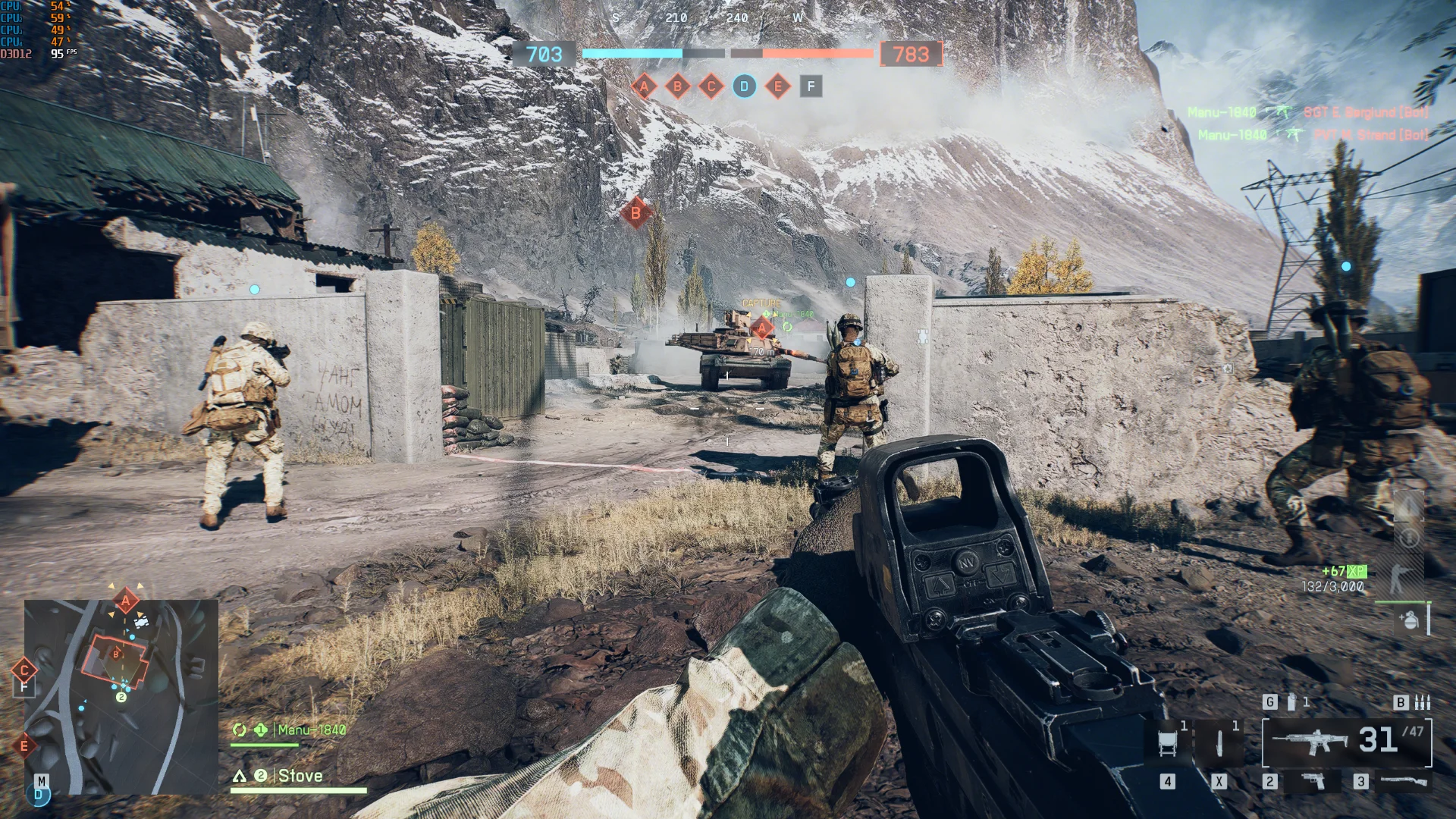}
    \end{tabular}
    \caption{\textbf{Environments used as research testbeds.} For this study, we use two popular AAA games to showcase the challenges in applying RL for game AI. \textbf{Top} shows an in-game screenshot of \emph{EA SPORTS FC 25}, a realistic physics-based football simulation game. In this environment, we try to improve the \emph{positioning} system of the goalkeeper AI with RL. \textbf{Bottom} shows an in-game screenshot of \emph{Battlefield 6}, a team-based, large-scale, multiplayer-oriented first-person shooter AAA game. In this testbed, we try to improve the locomotion system of on-ground soldiers. }
    \label{fig:screenshots}
\end{figure}

Although RL has primarily been applied to game testing in production settings, this technology has the potential to enhance gameplay experiences. Technological advances in the video game industry have created increasingly complex and immersive game environments. However, the development of Artificial Intelligence (AI), that control non-player characters (NPCs), is still a critical element in the creative process that can break or elevate the game experience for the player. RL has the potential to enhance game AI by enabling the creation of more authentic, reliable, and immersive NPCs~\cite{sestini2025human, sestini2020deepcrawl}. We envision more applications of RL-augmented game AI in the future. However, for this to materialize, current research limitations are prohibitive to broad deployment across game genres~\cite{jacob2020s}. Therefore, in this vision paper, we propose a framework for training reinforcement learning models with a set of requirements in mind that are suited towards game AI and game development:

\minisection{Short Training Time -} games in active development can change on a daily basis. An RL-based game AI system must therefore keep pace with the development cycle, requiring time-efficient retraining;

\minisection{Controllability -} designers and developers require qualitative control over the final behavior of the agents, in order to accomplish their vision of the game experience;

\minisection{Modularity -} game AI systems are often structured modularly, as this provides developers with a high degree of controllability. Furthermore, hand-coded solutions for specific behaviors can still offer advantages over RL-based approaches. As a result, fully end-to-end systems based solely on RL are often viewed as suboptimal from a developer’s perspective. The RL system needs to be easily integrated into existing or future game AI systems, with a modular design;

\minisection{Maintainability -} as we will see in Section~\ref{sec:game_ai}, a common drawback of hand-coded systems is that, as games scale in complexity and scope, the underlying AI systems become increasingly difficult to manage and maintain together with the introduced changes. Any RL system developed for a game must therefore remain maintainable over time and adaptable to future updates and releases of the same game;

\minisection{Bug Detection and Fixing -} game AI systems need to be properly tested before they are delivered within commercial video games. After training agents with RL, designers require a systematic way to confirm that the game AI has the correct behavior, and to easily fix it in case of a problem;

\minisection{Authenticity -} game AI systems do not require superhuman agents, but rather agents that are suitable for a playful experience. It is relatively easy to train RL agents with superhuman performances, but we argue that it is much harder to train authentic agents fulfilling the creative vision of the game; and

\minisection{Runtime Inference Constraints - } while it is relatively easy to train well-performing RL agents with large neural networks with generous compute resources, the reality is rarely the same for model deployment. Games need to be developed with the lowest hardware bound in mind. The model needs to run either on-device for single-player or peer-to-peer multiplayer games, or on dedicated game servers for larger-scale multiplayer games. This involves not only model inference, but also observation computation and collection. Further, the computational budget for an RL-augmented system must be shared with the non-RL parts of that system.

We argue that these challenges are overlooked in the current academic literature, which either focuses on simple environments that do not capture the complexity of AAA games, proposes complex systems that require substantial computational resources, or focuses on superhuman performance. 

This paper first describes how the aforementioned challenges affect RL's applicability in game production settings by reporting experiments integrating RL-based game AI into popular commercial AAA video games. The games we use as testbeds are \emph{EA SPORTS FC 25} and \emph{Battlefield 6}. For the former -- a realistic football-based simulation -- the goal is to improve the goalkeeper AI's positioning system using RL. For the latter -- a first-person shooter video game -- the goal is to improve the quality of ground infantry using RL. Finally, we will present lessons learned and outline research directions to address the defined challenges and increase the adoption of RL-based game AI in the game industry.

\section{Traditional Hand-Coded Game AI}
\label{sec:game_ai}
In this section, we list a popular class of hand-coded AI methods for game development~\cite{yannakakis2025ai} that are useful for discussing the test cases we will see in Section~\ref{sec:experiments}. 

\minisection{Finite State Machines.} A Finite State Machine (FSM) is a programming pattern that models the behavior of NPCs as a graph of distinct states and transitions between these states given certain triggers. An FSM is composed of three components: a finite set of states that store information about a task; a number of transitions between states, which are triggered by specific events or conditions; and a set of actions that need to be executed within each state. The FSM's simplicity and low computational cost makes it a practical choice for NPCs' control within game production. However, large-scale FSMs can be very complex to design and maintain, and they offer limited room for dynamism, adaptability, and evolution~\cite{yannakakis2025ai}.

\minisection{Behavior Trees.} A Behavior Tree (BT) is a modular game AI programming pattern that has been adopted in robotics~\cite{kartasev2023improving}. A BT represents decision-making as a tree structure composed of control-flow and task nodes. The latter represent the low-level action the NPC executes, while the former can be of three types: \textit{sequence},  with which the BT runs all child task nodes in sequence; \textit{selector}, with which the BT  runs specific child task nodes depending on metrics; and \textit{decorator}, which adds complexity to a child behavior~\cite{yannakakis2025ai}. Their modularity and readability make BTs among the most widely used game AI techniques today. However, their dynamism and adaptability are limited, as they are static knowledge representations~\cite{yannakakis2025ai}.

\minisection{Goal-Oriented Action Planning.} Goal-Oriented Action Planning (GOAP) is an AI architecture in which agents dynamically construct sequences of actions to achieve a desired goal. GOAP uses a planner, typically based on A* search, that works backward from the goal state: it evaluates the difference between the current world state and the goal, then chains together actions whose preconditions and effects bridge that gap. Each action in the system declares what world state it requires to execute and how it modifies the world state upon completion. The planner searches through these action combinations to produce a valid, cost-optimal plan at runtime. GOAP systems are limited by their reliance on engineered representations, which can become complex as the scale of the game increases; moreover, combinatorial search over the action space can become prohibitively expensive as the number of actions and world-state variables increases~\cite{jeff2003applying}.


These AI techniques rely on established path-finding algorithms, combined with Navigation Meshes (NavMeshes), which represent traversable surfaces within a game level and impose movement limitations not experienced by human players. Although ML-based game AI can mitigate the aforementioned problems, these handcrafted systems do have value for authoring NPCs' behavior, and we argue that game developers should leverage them to create compelling game AI. The use of ML-based agents should not be seen as a means of replacing hand-coded AI, but rather as a complement to these systems. To showcase examples of where and why these systems can fall short and how ML-based agents can complement them, in the next section, we will describe the AAA video games we used in this study and their AI systems.

\section{Case Studies}
\label{sec:environments}
In this section, we describe the two commercial video games we use as testbeds to showcase the challenges in applying RL-based game AI to AAA games. The games are \emph{EA SPORTS FC 25} and  \emph{Battlefield 6}.

\minisection{EA SPORTS FC 25.} The game is part of the \textit{EA SPORTS FC} series, with new entries released every year. 
Figure~\ref{fig:screenshots} top shows a screenshot of the game. The game is a physics-based football simulation where players compete against other humans and in-game AI. In this paper, we study the case of augmenting the goalkeeper's game AI with RL. The system is an FSM that comprises different states, e.g. \emph{saving} and \emph{positioning}. Based on the situation, the goalkeeper AI switches to the relevant state; for instance, when entering a shot situation, it switches from the \emph{positioning} state to the \emph{saving} state. The \emph{positioning} state -- the main state telling the goalkeeper where to move in most of the situations -- itself is another FSM with different states, based on the specific scenario, e.g. closing the angle between a potential striker and the goal. Although the current \emph{positioning} provides overall good behavior for the goalkeeper, it suffers from three problems: believability, maintainability, and overall performance. Using the hand-coded system, the sudden switch between states is evident, making the goalkeeper's movement less realistic. To cover all possible low-level cases, developers had to implement a large, complex FSM that is hard to expand and maintain. 

\minisection{Battlefield 6.} \emph{Battlefield 6} is a team-based, large-scale, multiplayer-oriented first-person shooter AAA game. Figure~\ref{fig:screenshots} bottom shows a screenshot of the game. This game has a multi-modal gameplay structure where the player tries to defeat enemies either as on-the-ground soldiers, in ground vehicles, or in aircrafts; in this paper, we explore soldiers as our use case. The soldiers' game AI is a mix of BT and GOAP that work together to deliver overall good performance. However, soldiers controlled by this system can exhibit unrealistic behavior, especially in locomotion -- i.e. the actual positioning and movement of the soldier in the environment. For example, the complexity of their AI system and its reliance on static representations (e.g., NavMesh) make it hard to create a believable locomotion system, leading soldiers to often follow paths rigidly and without variation to reach a target position, or to fail to consider and react to the environment around them. 

Although RL can mitigate the problems we mentioned when applied to the game AI in the testbed games, its application to large-scale, complex systems is not straightforward. An RL-based system integrated into such complex games must address the challenges described in Section~\ref{sec:introduction}, and we argue that current RL research is insufficient to fully address them. 

\section{Reinforcement Learning for Game AI}
\label{sec:experiments}
In this section, we use the AAA game environments mentioned in Section~\ref{sec:environments}, on the one hand, to showcase examples of how we can improve specific game AI systems with RL, and on the other, to describe the task specifications and challenges.

\subsection{EA SPORTS FC 25}
\label{sec:experiments_fc}
For this case, we are interested in replacing the goalkeeper's positioning system in its game AI.

\minisection{Task Specifications.} RL offers the potential to mitigate the three problems we identified in the hand-coded positioning system (see Section~\ref{sec:environments}): believability, maintainability, and overall performance. The behavior can be explained by a reward function rather than complex state interactions that an FSM would require, and this function can be defined by domain experts (e.g. professional goalkeepers) who are not skilled coders. The behavior is specified with a higher level of abstraction than coding, rewarding the agent for relevant micro behaviors and letting the algorithm figure out how to solve the task in both optimal and emergent fashion. This process can make it easier to create believable behaviors than by using FSMs. Moreover, in the case of RL, updating the agent's behavior means updating the reward function (and potentially the training scenarios, as we will see later) and restarting training. In the case of FSMs, updating a behavior may require a deep understanding of the existing system and potentially introduce changes that could break the FSM.

Here we list the requirements that the RL-based solution needs to satisfy:
\begin{itemize}
    \item \textit{Authenticity}: the main goal of this experiment is to have a more believable and human-like agent, compared to the existing hand-coded agent;
    \item \textit{Short Training Time}: modern game development follows a highly iterative process, where gameplay systems are regularly updated and refined. An RL training pipeline must therefore keep pace with frequent updates, requiring efficient retraining procedures. Our goal is to enable overnight training;
    \item \textit{Modularity}: the RL system needs to be integrated into an existing game AI logic, as we are improving only a small part of the entire system (i.e. the positioning system);
    \item \textit{Bug Detection and Fixing}: designers and developers need to be able to easily and quickly change the behavior of the agent in case they need to fix it; and
    \item \textit{Runtime Inference Constraints}: the model needs to run on-device, potentially in low-end machines such as video game consoles.
\end{itemize}

\begin{figure}
    \centering
    \begin{tabular}{cc}
         \includegraphics[width=0.25\textwidth]{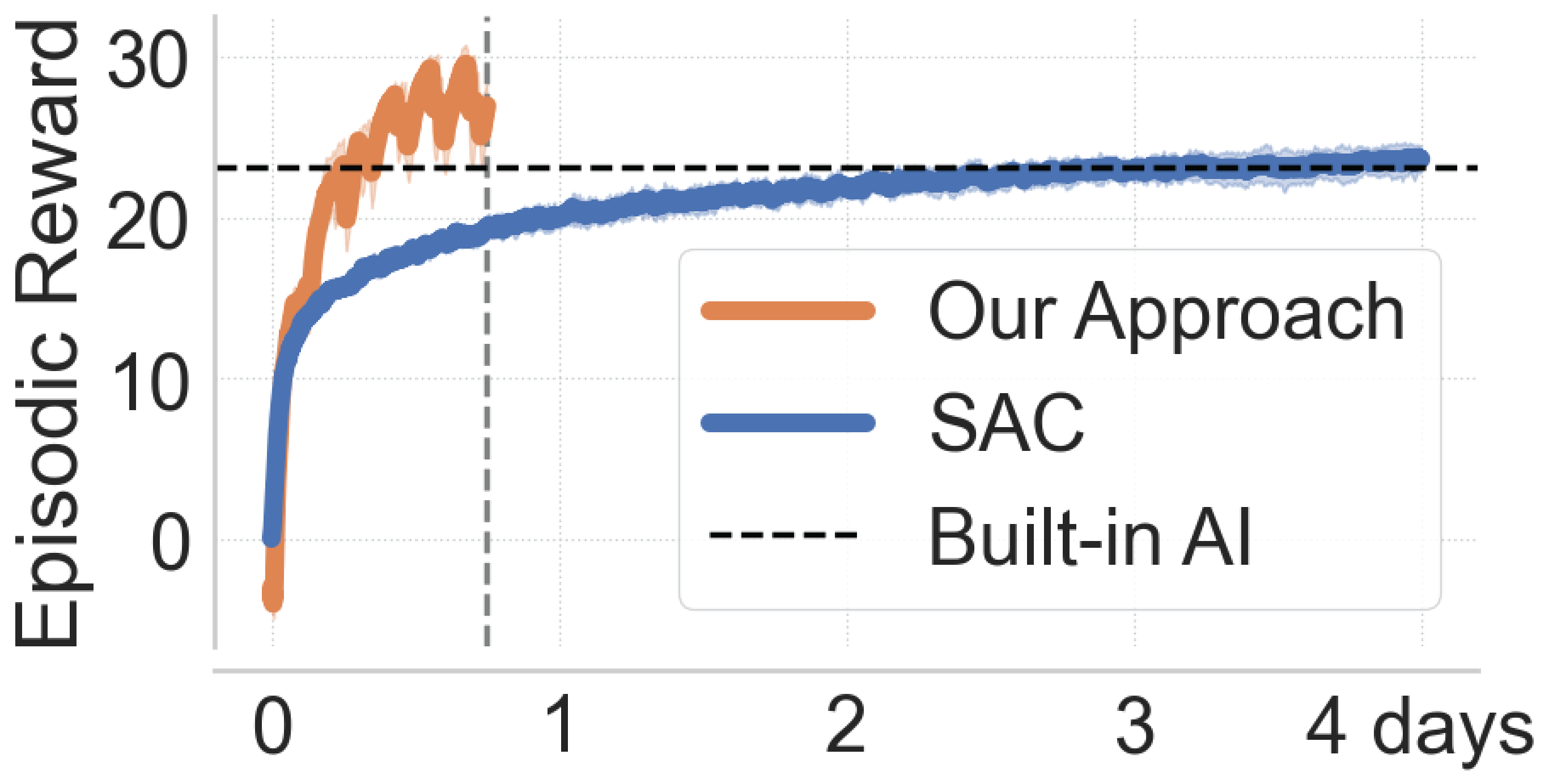}     &
         \includegraphics[width=0.25\textwidth]{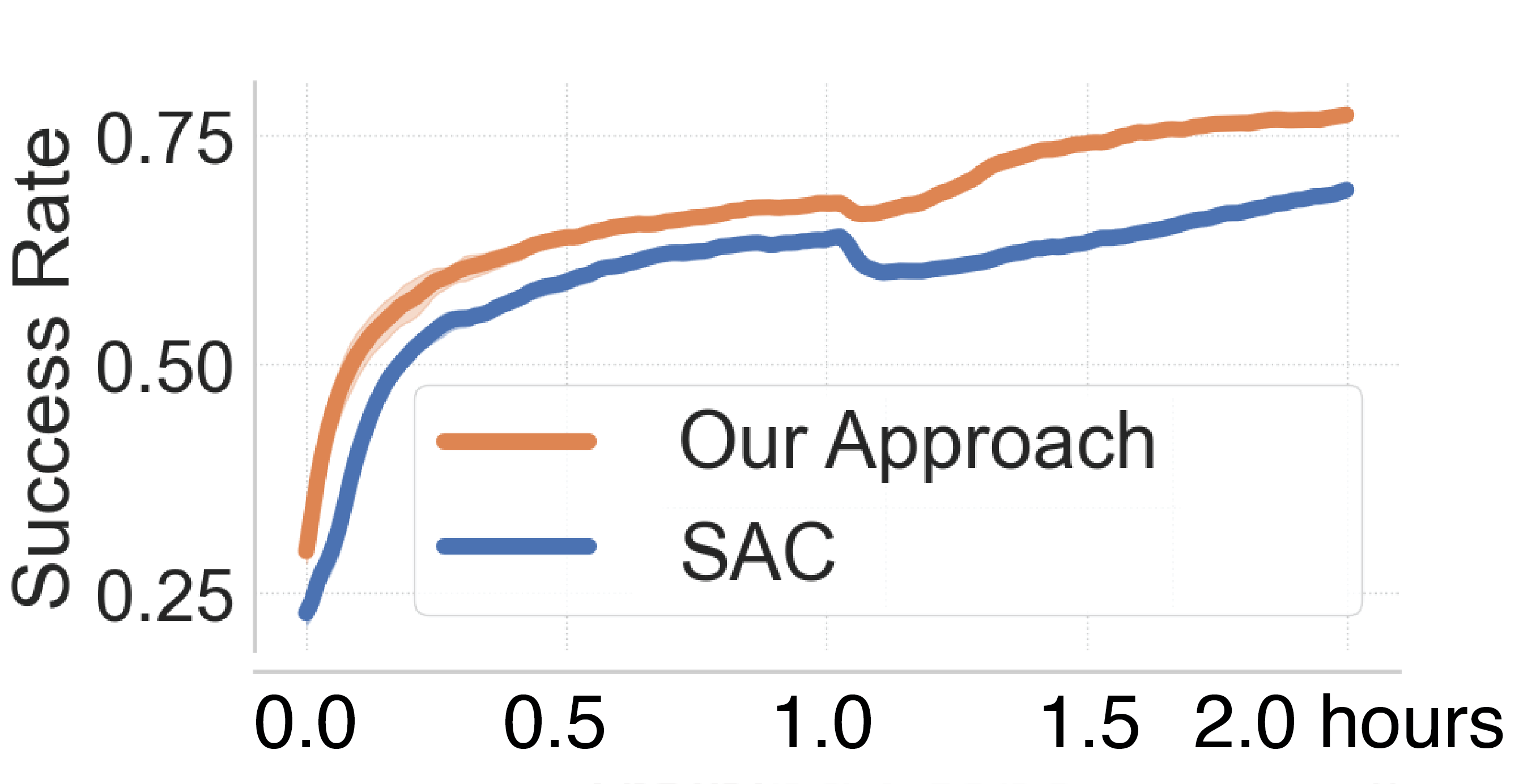}      
    \end{tabular}
    \caption{\textbf{Training performance using our approach in \textit{EA SPORTS FC 25}}. \textbf{Left} shows a comparison between an agent trained with standard SAC and our modified variant designed to satisfy the \textit{Short Training Time} requirement. The dotted line indicates the performance of the built-in hand-coded AI that the RL agent aims to augment. Through targeted modifications, the proposed approach improves sample efficiency, enabling effective overnight training. \textbf{Right} shows a comparison between standard SAC fine-tuning and our proposed fine-tuning strategy in a scenario where designers identified undesirable behavior. Our method provides a more effective alternative to conventional SAC fine-tuning, enabling targeted behavioral corrections in approximately 2 hours.}
    \label{fig:fc_exps}
\end{figure}

\minisection{Algorithm Choice.} One of the biggest limitations of RL from a game production perspective is its relatively long training time. Games under development are often computationally expensive to simulate, inherently unstable, and prone to crashes. At the same time, the \textit{modularity} requirement forces us to use the real game as a training environment, since the agent needs to interact with other systems in the game. For \textit{EA SPORTS FC 25}, we train the agent using a low-resolution configuration of the game. This setup removes non-essential graphical enhancements during training and enables unlocked frame rates, achieving up to 120 frames per second on a standard development machine (e.g. equipped with an NVIDIA RTX 4090 GPU). On the same machine, we can run five game instances in parallel during a training session. The agent executes one action every five frames, resulting in an overall throughput of approximately 120 samples per second across all instances. Compared to modern RL environments and frameworks, which can reach thousands of steps per second~\cite{suarez2025pufferlib}, this represents a relatively low data collection rate. Given these constraints, we selected the Soft Actor-Critic (SAC) algorithm~\cite{haarnoja2018soft}, as it is among the most sample-efficient RL methods. However, a single SAC training session still required between two and four days, initially. For context, the training had to be achievable \textit{at most overnight} to allow a developer to start a new training session at the end of the workday and obtain a newly trained agent by the following day. As described in prior work~\cite{sestini2025human}, to address this we employ several advanced techniques -- such as a high update-to-data ratio with network resets, the use of pre-collected offline data, and scenario-based training -- to reduce the duration of a SAC training session from $4$ days to approximately $12$ hours. Figure~\ref{fig:fc_exps} left shows the difference in training performance between standard SAC and the modified variant used in this experiment. Although these relatively simple modifications substantially reduce training time, we argue that further research in this direction is necessary, as we will discuss in Section~\ref{sec:research_directions}.

Regarding the neural network architecture, the \textit{runtime inference constraints} requirement requires us to use a compact, computationally efficient model. In this game, we have a strict budget of \SI{200}{\micro\second} per inference call. By inference call, we refer to the complete pipeline of retrieving observations from the game engine, executing the model's forward pass, and returning the selected action. The strict time constraint for inference stems from the \textit{modularity} requirement. The goalkeeper's AI system is implemented as a collection of interacting subsystems, each operating under its own CPU and GPU time allocation. We adopt a 5-layer Multi-Layer Perceptron (MLP) with SiLU activations and layer normalization, where each fully connected layer has $256$ hidden units. The resulting network contains approximately 300,000 parameters and achieves a total inference time of \SI{170}{\micro\second} in the lowest-end configuration we tested, remaining within the allocated budget. Rather than simply scaling model size, we argue that RL research should prioritize small, sample-efficient architectures tailored to real-time deployment, thus enabling production-ready RL solutions.

To minimize friction in integrating these approaches into development tools and pipelines, an RL-based game AI system needs to provide a quick, reliable way to iterate on the trained agent. This process is useful when the model is released alongside the finished game. Player-discovered exploits, in which specific behavioral weaknesses are systematically leveraged to score goals, can compromise the integrity of the gameplay experience. Quick and rapid mitigation of such behaviors is therefore essential. In this scenario, the training time budget requirement is significantly lower than for a full training cycle -- approximately 4 hours. We make use of our previous work~\cite{sestini2025human} and combine scenario-based learning with the Replay across Experiments (RaE) technique~\cite{tirumala2023replay}. This approach enables targeted fine-tuning of the RL agent within 2 to 4 hours, depending on the exploit's complexity, thereby satisfying the production constraint. As with the main training pipeline, relatively simple modifications bring RL-based solutions closer to a production-ready tool. However, this strategy has diminishing returns: repeated fine-tuning increases the risk of catastrophic forgetting. Figure~\ref{fig:fc_exps} right compares standard SAC fine-tuning with our proposed approach.

The resulting agent effectively addresses the three challenges identified earlier. The new positioning system exhibits an overall behavior that playtesters perceive as more believable, authentic, and human-like. Moreover, behavioral adjustments can be implemented easily by developers and designers. As a side note, with the new system, the goalkeeper achieves a 10\% higher save ratio compared to the previous hand-coded solution.  


\addtolength{\tabcolsep}{-0.7em}
\setlength{\tabcolsep}{6pt}

\subsection{Battlefield 6}
\label{sec:experiments_bf}
As mentioned in Section~\ref{sec:environments}, we are interested in augmenting the underlying game AI system with improved \textit{locomotion} behavior using RL.

\minisection{Task Specification.} In this task, our objective is less focused on designing a rich, dense reward function and more concerned with addressing limitations of the existing locomotion game AI system, such as rigid pathfinding and limited contextual awareness. For locomotion, the underlying BT relies on a pathfinding algorithm and navigation meshes to traverse to a given destination. With RL, instead of relying on state discretization, the model can learn from continuous-state observations and directly output continuous movements. While effective, such solutions can result in behavior that appears artificial. 

To be practically integrated into the current development cycle, a potential RL approach must satisfy several requirements, some of which overlap with those discussed in the previous use case:
\begin{itemize}
\item \textit{Short Training Time}: as for \textit{EA SPORTS FC 25}, the game will continuously change during development, requiring the RL training framework to allow for rapid adaptation to frequent gameplay and system changes;
\item \textit{Modularity}: to preserve the existing BT structure, the \textit{locomotion} components must be implemented as separate agents, replacing the corresponding leaf node within the BT and must be able to interact with other hand-coded nodes;
\item \textit{Runtime Inference Constraints}: in this setting, the primary computational bottleneck is observation collection rather than model inference time; and
\item \textit{Authenticity}: the main objective is not to maximize quantitative performance, but to generate behavior that appears more authentic and believable compared to the current hand-coded implementation.
\end{itemize}


\addtolength{\tabcolsep}{-0.3em}
\begin{figure}
    \centering
    \begin{tabular}{cc}
         \includegraphics[width=0.20\textwidth]{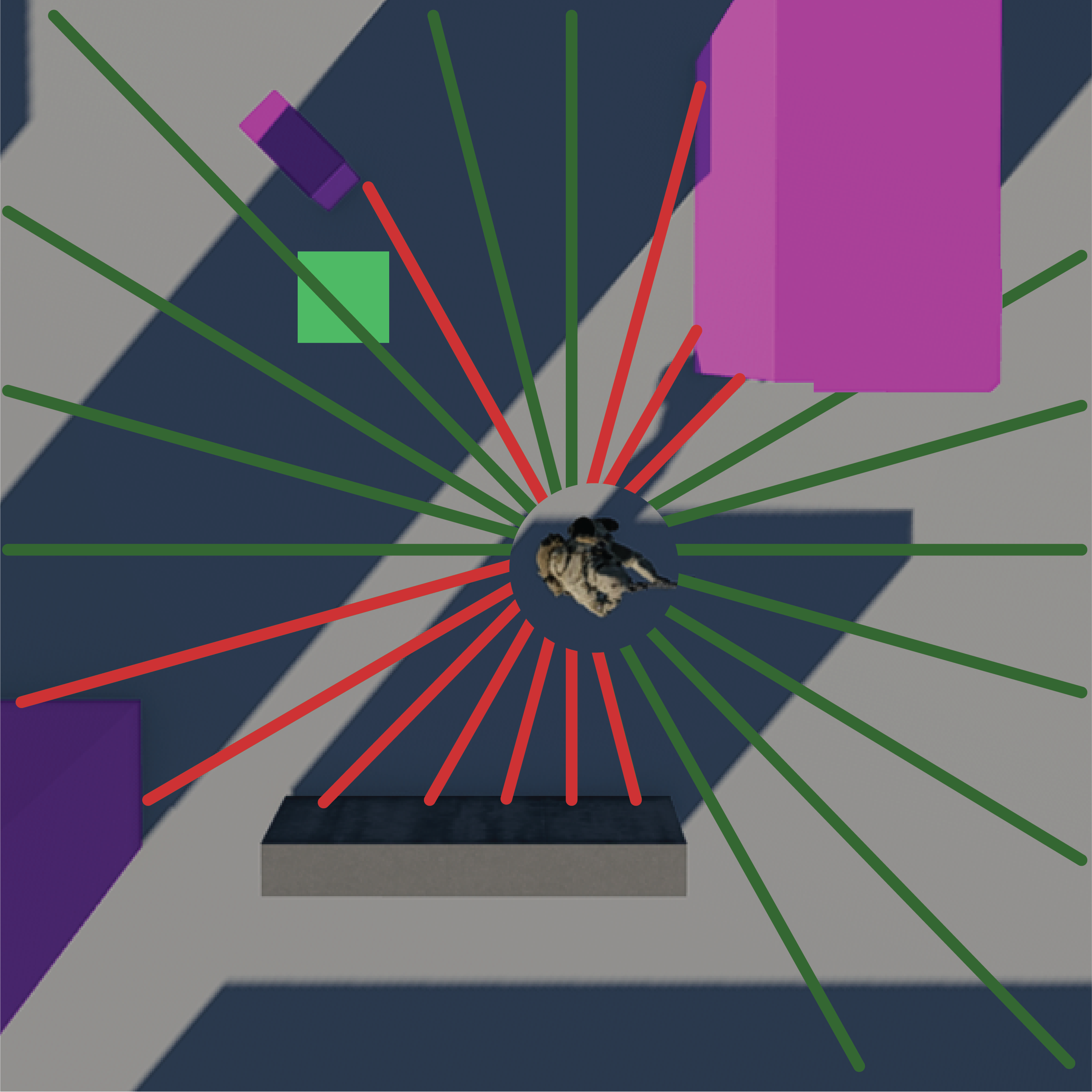}     &
         \includegraphics[width=0.20\textwidth]{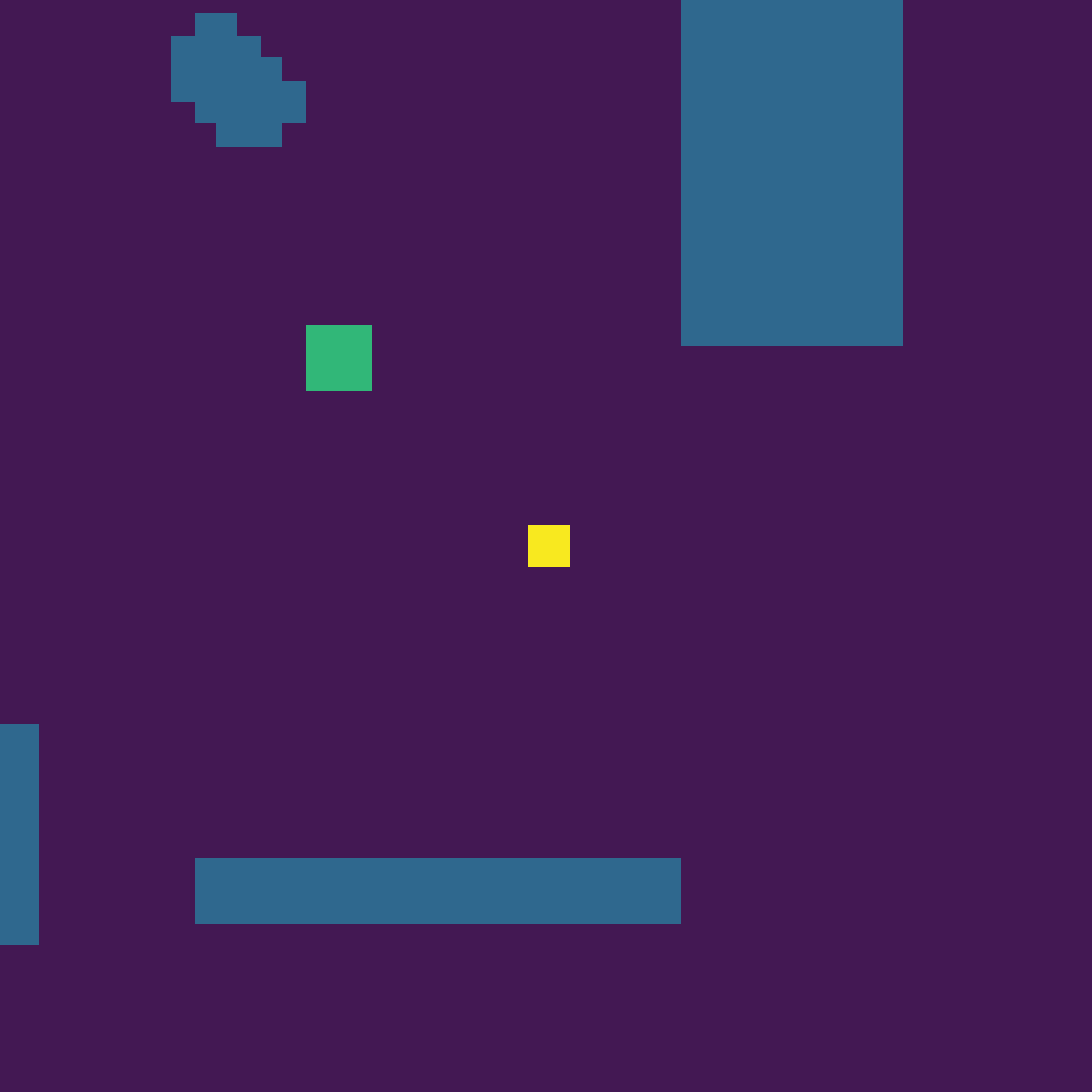}       
    \end{tabular}
    \caption{\textbf{Modes of perception explored in \textit{Battlefield 6}}. \textbf{Left} Raycast fan of $24$ rays with high enough density to allow detection of all obstacles within a \SI{10}{\meter} radius of the agent. \textbf{Right} Occupancy map of size $50 \times 50$. Each pixel color represents either: agent (yellow), terrain (purple), obstacle (blue) or target waypoint (green). When the waypoint is out of range, it is mapped to the border of the occupancy map as a directional indicator.}
    \label{fig:bf_perception}
\end{figure}
\setlength{\tabcolsep}{6pt}

\addtolength{\tabcolsep}{-1.4em}
\begin{figure}
    \centering
    \begin{tabular}{cc}
         \includegraphics[width=0.27\textwidth]{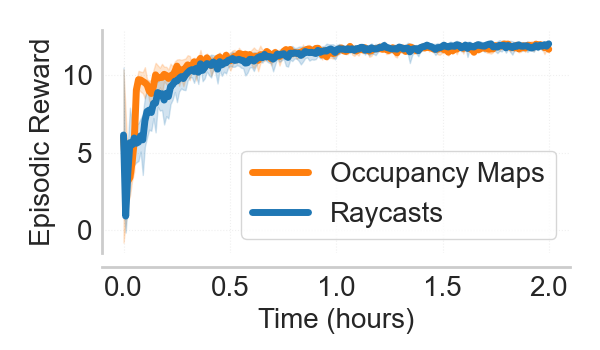}     &
         \includegraphics[width=0.27\textwidth]{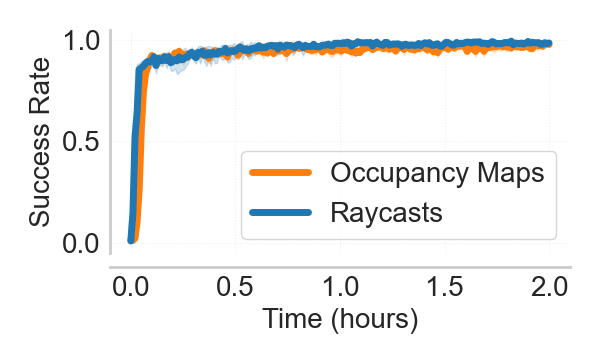}       \\
    \end{tabular}
    \caption{\textbf{Comparison of agent training with raycasts (blue) and occupancy maps (orange) in \textit{Battlefield 6}}. \textbf{Left} Average episodic reward. \textbf{Right} Average rate over time where the agent successfully finds the target waypoint before a set step limit of $500$. As seen in the graphs, both approaches lead to comparable performance. However, occupancy maps are computationally faster to run by a factor of $\approx 2.0$, making the approach more attractive in a game production setting.}
    \label{fig:bf_experiments}
\end{figure}
\setlength{\tabcolsep}{6pt}

\begin{figure}
    \centering
    \begin{tabular}{c}
         \includegraphics[width=0.75\linewidth]{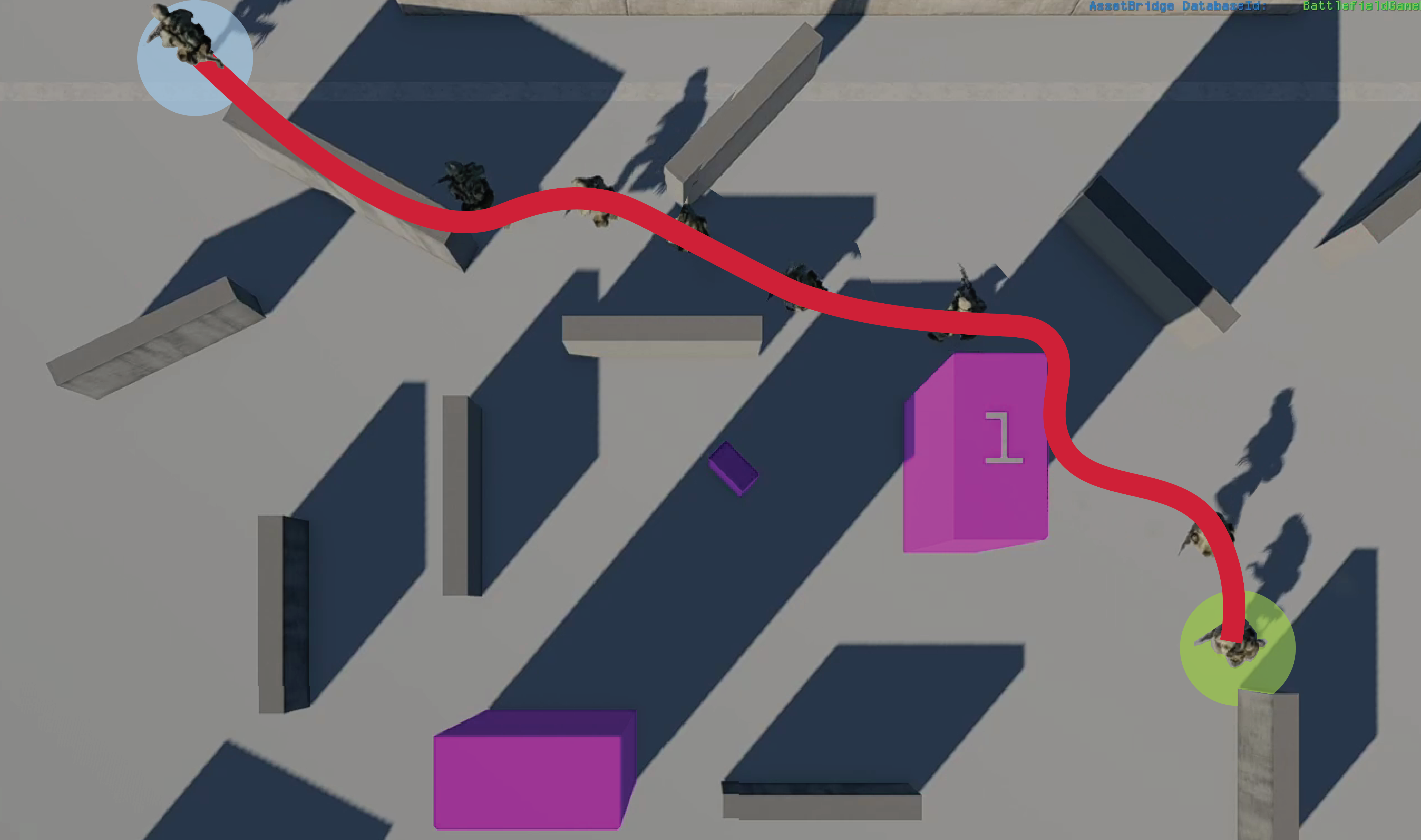} \\
         \includegraphics[width=0.75\linewidth]{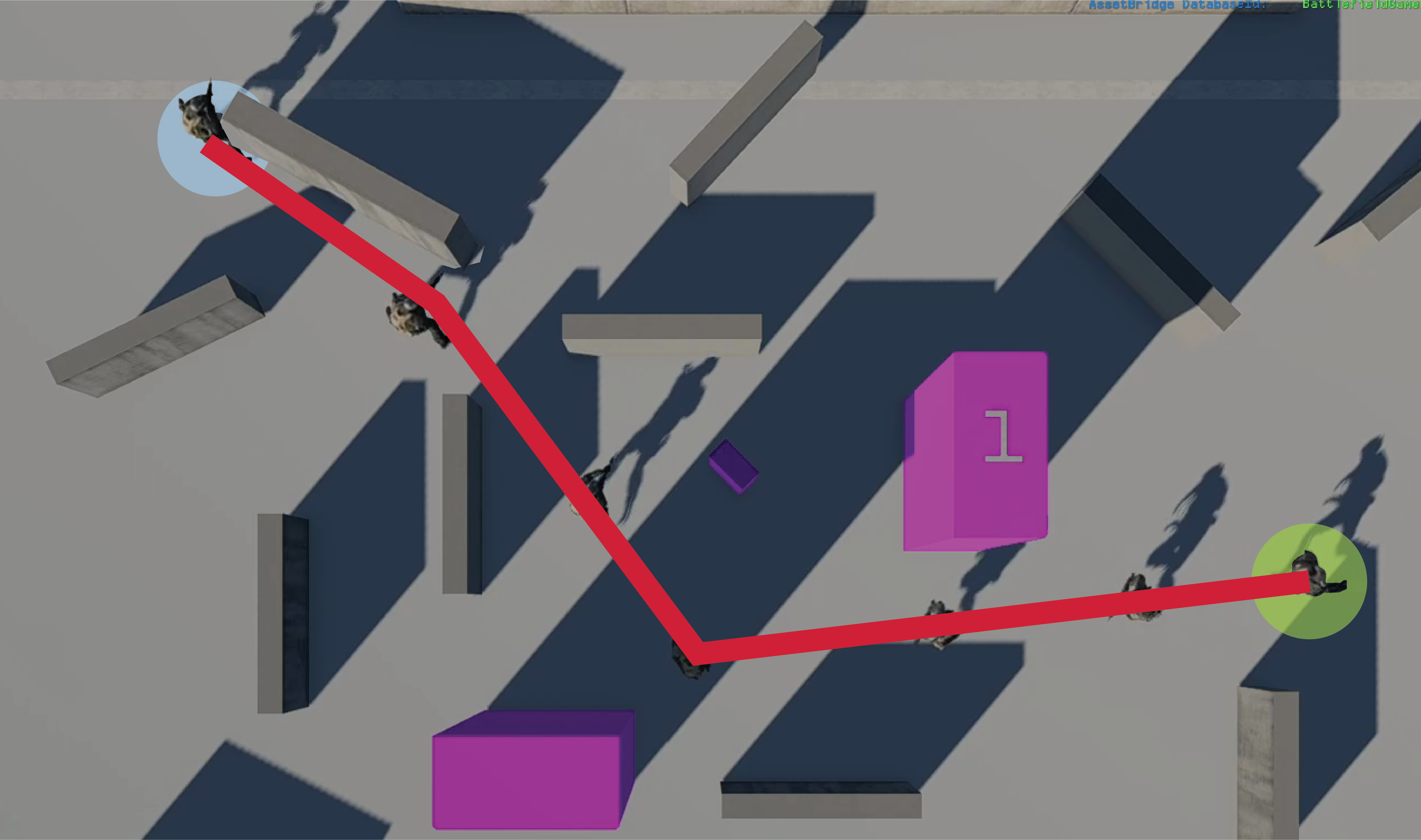}
    \end{tabular}
    \caption{\textbf{Comparison of locomotion systems in \textit{Battlefield 6}.} \textbf{Top}: RL-augmented locomotion system. \textbf{Bottom}: NavMesh-based game AI. The RL-augmented agent exhibits smoother, more natural trajectories that resemble human player behavior. In contrast, the hand-coded system -- mainly due to the discretization of the NavMesh representation -- produces more rigid and less realistic movement patterns.}    
    \label{fig:qualitative_results_bf}
\end{figure}

\minisection{Algorithm Choice.}  In contrast to \textit{EA SPORTS FC 25}, sample efficiency is not the main challenge in this setting. For this particular game, we can instantiate multiple environments on the same machine, each containing several agents running in parallel. Furthermore, training can be conducted using a \textit{headless} dedicated server version of the game, without a graphical interface, increasing simulation throughput. In total, this setup allows up to 240 agents to run concurrently. Given these conditions, we prioritize a simpler, more stable algorithm such as Proximal Policy Optimization (PPO), which is less computationally intensive than SAC. Training time is approximately $2$ hours.

We are interested in a simple task: navigating the environment from a random starting position to a random target waypoint while avoiding obstacles. The agent is rewarded monotonically for reducing its distance to the target waypoint, and is penalized for colliding with obstacles. Figure~\ref{fig:qualitative_results_bf} shows an example of the training environment. Contrary to many state-of-the-art approaches used in literature, which rely on raycasting techniques~\cite{justesen2025human, alonso2020deep} or first-person vision-based agents~\cite{pearce2022counter}, the \textit{Runtime Inference Constraints} requirement prevents us from adopting such solutions. Extracting these representations requires significant computational resources, making observation collection -- rather than model inference -- the main bottleneck in our setting. Instead, we leverage data structures already available within the game engine, specifically heightmaps, which we repurpose as occupancy maps centered around the agent. At game boot-up, we leverage the initial step of the NavMesh generation process to cache a level-wide representation of where non-traversable areas and obstacles are. At runtime, we query this cache using the agent's current in-world position for a local region. Before constructing the final observation for the agent, we construct the occupancy map with $4$ categorical pixel value types: the agent, terrain, obstacles, and finally the target waypoint. 

To signify the cost difference between occupancy maps and raycasts, we run a baseline experiment utilizing $24$ raycasts in a $360^{\circ}$ fan around the agent~\cite{justesen2025human}. Figure~\ref{fig:bf_perception} showcases examples of both the raycast fan and occupancy map respectively. The resulting neural network architecture is composed of two input paths: an MLP of three linear layers, sized $64$, $128$, and $64$ respectively for auxiliary game features, and either an additional three-layer network of the same shape when using a raycast-based perception, or a $5$ layer convolution-based encoder when using the aforementioned occupancy map,  
all with a kernel size of $3 \times 3$, stride of $2$ and no padding, besides the first layer responsible for the pixel type embedding which utilizes a kernel size of $1 \times 1$. Figure~\ref{fig:bf_experiments} compares training performance when using raycast-based observations versus occupancy map representations. While both approaches achieve comparable overall performance, occupancy maps are significantly less computationally expensive. Comparing the average wall-clock time of running the aforementioned $24$ raycasts with building the occupancy map, the former takes about \SI{27}{\micro\second} while the latter only \SI{14}{\micro\second}, amounting to a $\approx 2.0 \times$ speed-up even with a conservative number of raycasts. Nevertheless, the occupancy map representation remains suboptimal. It does not capture several important aspects of game AI, such as multi-layered environments, vertical structures, irregular terrain, dynamic obstacles, and destruction. We therefore argue that further research is needed to develop computationally efficient yet expressive game-state representations that better balance realism, scalability, and deployment constraints in production environments. Figure~\ref{fig:qualitative_results_bf} shows the qualitative difference between the human-like locomotion of the RL agent, compared to the existing NavMesh-reliant system, showing that the RL-based agent better satisfies the \textit{Authenticity} requirement.

As we mentioned in the \textit{Modularity} requirement, the RL-based \textit{locomotion} system needs to be integrated with other existing nodes in the game AI bots. To illustrate, we run an experiment in a testing environment where an RL-augmented agent faces a hand-coded opponent in a 1-on-1 scenario. To get a high success rate, the RL-based \textit{locomotion} node needs to interact with other systems, such as the \textit{aiming and shooting} behavior. We run 20 testing episodes, with the RL-augmented agent winning 11 episodes. This shows that the RL-augmented locomotion has performance similar to the hand-coded game AI but exhibits more authentic behavior. Although recent work suggests that jointly training multiple leaves in an RL-augmented BT can yield improved performance~\cite{kartasev2023improving}, this experiment demonstrates that even this simpler approach -- i.e. training policies independently and integrating them within the BT -- meets the \textit{Modularity} requirement while maintaining satisfactory behavior.

\section{Future Research Directions}
\label{sec:research_directions}
In Section~\ref{sec:experiments}, we showed how RL-augmented game AI can improve the overall performance of standard techniques. On the other hand, RL integration into game AI comes with a set of requirements that are often overlooked in the current literature, but that we argue are fundamental to enabling the mass adoption of RL in game AI. In this section, we propose research opportunities that we believe will be essential for driving research and, therefore, industry forward.
 
\minisection{Designers-in-the-Loop.} To satisfy requirements such as \textit{Controllability} and \textit{Maintainability}, an RL-based game AI system must prioritize designers as primary users. Designers require qualitative control over the agent’s final behavior in order to express creative intent and ensure alignment with the overall gameplay vision. For this reason, they need to have an active role within the learning framework. The approach discussed in Section~\ref{sec:experiments_fc}, originally introduced in our previous work~\cite{sestini2025human}, represents a step in this direction. While human-in-the-loop learning has demonstrated promising results in domains such as large language models~\cite{ouyang2022training} and robotics~\cite{marta2025flora} -- primarily to improve safety and alignment -- applications in games have largely focused on improving task performance or sample efficiency. \citet{sestini2023towards} propose an interactive method for creating playtesting agents; however, playtesting agents operate under different requirements and goals than production game AI agents~\cite{gillberg2023technical}. Similarly, \citet{zhao2020winning} adopt a behavioral cloning–like approach in which agents learn policies directly from designer demonstrations, relying only on explicit demonstrations as supervision. To the best of our knowledge, existing work does not address the problem of providing designers with an authoring tool for shaping qualitative behaviors in production game AI. Furthermore, the question of how designers should interact with the training process -- and what representation of feedback (e.g. demonstrations, preference comparisons, or dense reward shaping) is most effective -- remains unexplored.

\minisection{Efficient Networks and Fast Training.} During a video game development cycle, the game changes on a daily basis -- for example, through changes in assets, mechanics, or gameplay dynamics. Such updates may require retraining the same agent from scratch multiple times. To remain practical, the training process must therefore keep pace with development, requiring fast turnaround times. Recent research has gone into two main directions to accelerate RL training. On one hand, projects such as PufferLib~\cite{suarez2025pufferlib} propose high-speed, lower-fidelity simulators combined with domain randomization, enabling agents to be trained efficiently and transferred to the target environment (i.e. the full game). On the other hand, a growing body of work focuses on improving sample and computational efficiency directly within the target environment~\cite{schwarzer2023bigger, romeo2025speq}. Given the complexity of AAA video games -- and in light of the \textit{Modularity} requirement -- we argue that the latter approach is more viable in this production setting. Games are complex systems, and constructing simplified simulators, particularly when the goal is to augment existing AI components, can be expensive or impractical. The \textit{Runtime Inference Constraints} limit the size of the network, with game studio preferring small and efficient architectures tailored for real-time deployment. For example, one can think of using an asymmetric actor-critic architecture~\cite{mastikhina2025optimistic} where the critic has a larger size and expressivity~\cite{nauman2024bigger} than the actor. Finally, such efficiency come with further software and hardware dependencies that define the deployment environment. Can the model run on GPU, or only on CPU? Which framework (e.g. ONNX, TensorRT) is needed to deploy the model?

\minisection{Fine-Tuning and Correcting Behaviors.} Modern games are constantly evolving live services for the duration of their lifecycle. Even after release, patches may introduce gameplay changes -- for instance, to address newly discovered player exploits. An RL-based game AI system must therefore be able to adapt rapidly to such updates. However, fine-tuning RL agents -- particularly when qualitative aspects of behavior must be modified -- remains a challenging problem. RL policies tend to overfit to the training environment, and once fine-tuned, they risk catastrophic forgetting of the pre-trained behaviors~\cite{wolczyk2024fine, zhang2018study}. The \textit{Modularity} requirement directly addresses this issue. Decomposing the overall behavior into smaller policies responsible for specific tasks can simplify iterative updates, as we may need to fine-tune only a specific sub-policy, rather than the entire system. We argue that systematic methods for behavioral fine-tuning in RL remain underexplored. While recent work on open-ended and continual learning~\cite{faldor2024omni, dohare2024loss} demonstrates that agents can adapt over long time horizons, it is still unclear how such approaches can be translated into practical, production-oriented workflows, such as those required in game development.

\minisection{Modularity versus End-to-End.} Recent RL research has largely focused on developing end-to-end models capable of handling all aspects of game AI within a single unified architecture~\cite{wurman2022outracing, vinyals2019grandmaster}. While such approaches have demonstrated impressive results in controlled settings, the practical constraints of game development often make end-to-end solutions impractical for production use, due to high research demands and substantial training times. In contrast, established systems such as the cited FSMs and BTs have been widely adopted in the industry due to their modularity and interpretability. Although these systems have limitations, they provide reliable and efficient solutions. Rather than replacing these frameworks entirely, RL-based game AI should focus on augmenting existing systems to overcome their specific shortcomings. The examples presented in Section~\ref{sec:experiments} demonstrate that both FSMs and BTs can be augmented with RL components to produce qualitatively improved behaviors. Moreover, recent work on RL-augmented BTs~\cite{kartasev2023improving} suggests that such modular integrations are also beneficial in other domains, including robotics. The problem of jointly training modular policies within a partially hand-authored control architecture remains unsolved. Moreover, how to switch policies in a BT, and the best switching handover, remain open challenges.

\minisection{Perception.} As discussed in Section~\ref{sec:experiments_bf}, \textit{Runtime Inference Constraints} limit the applicability of RL-based game AI not only due to the use of large neural networks, but also because of the cost of state extraction. Standard state representations used by modern approaches~\cite{pearce2022counter, alonso2020deep, justesen2025human} often require significant computational resources, which are often unavailable in commercial video games that must run across a wide range of hardware platforms. 3D spatial understanding remains fundamental for embodied agents in games. Our experiments with occupancy maps demonstrate that it is possible to achieve performance comparable to state-of-the-art methods using more computationally efficient representations. However, occupancy maps provide only a partial representation of the surrounding 3D environment. We argue that further research is needed to develop computationally efficient yet expressive 3D representations, as explored in recent work such by \citet{ying2024efficient}.

\minisection{Authenticity.} Recent successes of RL in video games have shown impressive results, including superhuman agents capable of defeating professional players in complex, skill-intensive environments~\cite{wurman2022outracing, vinyals2019grandmaster}. However, achieving superhuman performance is not the primary goal of RL-based game AI in production contexts. From a player’s perspective, competing against a superhuman agent can be frustrating and may detract from the intended gameplay experience. Instead, RL-augmented game AI should contribute meaningfully to game design by producing behaviors that are authentic and believable. For example, in a game such as \textit{EA SPORTS FC 25}, a goalkeeper should behave like a human professional goalkeeper; similarly, in a game such as \textit{Battlefield 6}, an infantry soldier should exhibit credible tactical behavior. Translating such qualitative design goals into explicit mathematical reward functions, however, is often difficult and unintuitive. Large-scale behavioral cloning approaches~\cite{yue2026scaling, magne2026nitrogen, pearce2022counter} and methods based on inverse RL~\cite{ahlberg2023generating, sestini2021policy} have been explored to generate more human-like behaviors in games. Behavioral cloning typically requires extensive datasets of human demonstrations, which are difficult to collect, particularly during active development. In contrast, a production-oriented system should ideally learn from a limited number of demonstrations provided directly by designers. However, in case the goal is to provide player-like behavior, large-scale behavioral cloning is indeed a viable research direction~\cite{pearce2022counter}. Inverse RL-based approaches can, in principle, operate with smaller datasets, but they often exhibit instabilities, which can limit their reliability in production environments~\cite{fu2017learning, sestini2021policy}. 

\minisection{Behavior Evaluation.} Within the video game development process, all features -- including game AI components -- must undergo proper testing and validation. However, evaluating RL-augmented game AI systems presents unique challenges, as such models often behave as black boxes, making them difficult to interpret and assess. Robust evaluation, including quantitative performance metrics and qualitative behavioral analysis, is essential to building trust among developers and designers. We argue that systematic evaluation and validation frameworks for RL-based game AI remain underexplored in the current literature, despite being a critical prerequisite for their mass adoption in commercial game development.

\section{Conclusions}
In this paper, we described the challenges that arise when RL is directly applied to game AI in AAA production environments, and we analyzed the technical challenges associated with meeting these requirements. Through our experiments, we demonstrated that these constraints can be addressed through targeted research and modifications of established RL algorithms. Our results suggest that RL offers a promising framework for augmenting rather than replacing traditional game AI techniques. However, for RL to achieve broader adoption in game production, improvements in usability, stability, controllability, and integration workflows are still needed. We have outlined several directions that we believe are relevant for advancing the adoption of RL-based game AI. 

In this work, we have focused on RL as one possible ML–based augmentation for game AI. However, alternative approaches such as large language model based systems also offer promising directions~\cite{justesen2019deep, gallotta2024large}. For example, one could imagine a large language model integrated within a BT architecture, where leaf nodes are implemented through either hand-authored logic or RL policies. We argue that RL currently represents one of the most mature and practically deployable techniques for augmenting existing game AI pipelines. We leave a systematic exploration of how other ML paradigms can be integrated into production-ready game AI systems for future research.

\bibliographystyle{IEEEtranN}
{\footnotesize \bibliography{references}}

@article{ying2024efficient,
  title={Efficient visibility approximation for game AI using neural omnidirectional distance fields},
  author={Ying, Zhi and Edwards, Nicholas and Kutuzov, Mikhail},
  journal={ACM on Computer Graphics and Interactive Techniques},
  year={2024},
}

@article{nauman2024bigger,
  title={Bigger, regularized, optimistic: scaling for compute and sample efficient continuous control},
  author={Nauman, Michal and Ostaszewski, Mateusz and Jankowski, Krzysztof and and others},
  journal={Advances in neural information processing systems},
  year={2024}
}

@article{mastikhina2025optimistic,
  title={Optimistic critics can empower small actors},
  author={Mastikhina, Olya and Sreenivas, Dhruv and Castro, Pablo Samuel},
  journal={arXiv preprint arXiv:2506.01016},
  year={2025}
}

@article{fu2017learning,
  title={Learning robust rewards with adversarial inverse reinforcement learning},
  author={Fu, Justin and Luo, Katie and Levine, Sergey},
  journal={arXiv preprint arXiv:1710.11248},
  year={2017}
}

@article{jeff2003applying,
  title={Applying goal-oriented action planning to games},
  author={Jeff, ORKIN},
  journal={AI game programming wisdom},
  year={2003}
}

@article{ouyang2022training,
  title={Training language models to follow instructions with human feedback},
  author={Ouyang, Long and Wu, Jeffrey and Jiang, Xu and Almeida, Diogo and Wainwright, Carroll and others},
  journal={Advances in neural information processing systems},
  year={2022}
}

@article{wolczyk2024fine,
  title={Fine-tuning reinforcement learning models is secretly a forgetting mitigation problem},
  author={Wo{\l}czyk, Maciej and Cupia{\l}, Bart{\l}omiej and Ostaszewski, Mateusz and Bortkiewicz, Micha{\l} et al.},
  journal={arXiv preprint arXiv:2402.02868},
  year={2024}
}

@article{zhang2018study,
  title={A study on overfitting in deep reinforcement learning},
  author={Zhang, Chiyuan and Vinyals, Oriol and Munos, Remi and Bengio, Samy},
  journal={arXiv preprint arXiv:1804.06893},
  year={2018}
}

@article{alonso2020deep,
  title={Deep reinforcement learning for navigation in AAA video games},
  author={Alonso, Eloi and Peter, Maxim and Goumard, David and Romoff, Joshua},
  journal={arXiv preprint arXiv:2011.04764},
  year={2020}
}

@article{vinyals2019grandmaster,
  title={Grandmaster level in StarCraft II using multi-agent reinforcement learning},
  author={Vinyals, Oriol and Babuschkin, Igor and Czarnecki, Wojciech M and Mathieu, Micha{\"e}l and Dudzik, Andrew and others},
  journal={Nature},
  year={2019},
}

@article{wurman2022outracing,
  title={Outracing champion Gran Turismo drivers with deep reinforcement learning},
  author={Wurman, Peter R and Barrett, Samuel and Kawamoto, Kenta and MacGlashan, James and Subramanian, Kaushik and Walsh, Thomas J and Capobianco, Roberto and Devlic, Alisa and Eckert, Franziska and Fuchs, Florian and others},
  journal={Nature},
  year={2022},
}

@article{berner2019dota,
  title={Dota 2 with large scale deep reinforcement learning},
  author={Berner, Christopher and Brockman, Greg and Chan, Brooke and others},
  journal={arXiv preprint arXiv:1912.06680},
  year={2019}
}

@inproceedings{bairamian2023minimax,
  title={Minimax Exploiter: A Data Efficient Approach for Competitive Self-Play},
  author={Bairamian, Daniel and Marcotte, Philippe and Romoff, Joshua and Robert, Gabriel and Nowrouzezahrai, Derek},
  booktitle={Proceedings of the 23rd International Conference on Autonomous Agents and Multiagent Systems},
  year={2024}
}

@article{wei2022honor,
  title={Honor of kings arena: an environment for generalization in competitive reinforcement learning},
  author={Wei, Hua and Chen, Jingxiao and Ji, Xiyang and Qin, Hongyang and Deng, Minwen and Li, Siqin and Wang, Liang and Zhang, Weinan and Yu, Yong and Linc, Liu and others},
  journal={Advances in Neural Information Processing Systems},
  year={2022}
}

@inproceedings{gillberg2023technical,
  title={Technical challenges of deploying reinforcement learning agents for game testing in aaa games},
  author={Gillberg, Jonas and Bergdahl, Joakim and Sestini, Alessandro and Eakins, Andrew and Gissl{\'e}n, Linus},
  year=2023,
  booktitle={IEEE Conference on Games (CoG)},
}

@inproceedings{sestini2023towards,
  title={Towards informed design and validation assistance in computer games using imitation learning},
  author={Sestini, Alessandro and Bergdahl, Joakim and Tollmar, Konrad and Bagdanov, Andrew D and Gissl{\'e}n, Linus},
  year=2023,
  booktitle={IEEE Conference on Games (CoG)},
}

@article{sestini2022automated,
  title={Automated gameplay testing and validation with curiosity-conditioned proximal trajectories},
  author={Sestini, Alessandro and Gissl{\'e}n, Linus and Bergdahl, Joakim and Tollmar, Konrad and Bagdanov, Andrew David},
  journal={IEEE Transactions on Games},
  year={2022},
}

@misc{modl,
  Author = "modl",
  Title  = "\emph{modl.ai}",
  Note   = "\url{https://modl.ai/}
           [Accessed: 2026]",
  year = 2026,
}

@misc{nunu,
  Author = "nunu",
  Title  = "\emph{nunu}",
  Note   = "\url{https://nunu.ai/}
           [Accessed: 2026]",
  year = 2026,
}

@inproceedings{sestini2025human,
  title={Human-Like Goalkeeping in a Realistic Football Simulation: a Sample-Efficient Reinforcement Learning Approach},
  author={Sestini, Alessandro and Bergdahl, Joakim and Barrette-LaPierre, Jean-Philippe and Fuchs, Florian and Chen, Brady and Jones, Michael and Gissl{\'e}n, Linus},
  booktitle={Reinforcement Learning Conference},
  year={2026}
}

@article{sestini2020deepcrawl,
  title={Deepcrawl: Deep reinforcement learning for turn-based strategy games},
  author={Sestini, Alessandro and Kuhnle, Alexander and Bagdanov, Andrew D},
  journal={arXiv preprint arXiv:2012.01914},
  year={2020}
}

@inproceedings{jacob2020s,
  title={“It’s unwieldy and it takes a lot of time”—challenges and opportunities for creating agents in commercial games},
  author={Jacob, Mikhail and Devlin, Sam and Hofmann, Katja},
  booktitle={AAAI Conference on Artificial Intelligence and Interactive Digital Entertainment},
  year={2020}
}

@incollection{yannakakis2025ai,
  title={AI Methods for Games},
  author={Yannakakis, Georgios N and Togelius, Julian},
  booktitle={Artificial Intelligence and Games},
  year={2025},
  publisher={Springer}
}

@inproceedings{kartasev2023improving,
  title={Improving the performance of backward chained behavior trees that use reinforcement learning},
  author={Kartasev, Mart and Saler, Justin and {\"O}gren, Petter},
  booktitle={2023 IEEE/RSJ International Conference on Intelligent Robots and Systems (IROS)},
  year={2023},
}

@article{haarnoja2018soft,
  title={Soft actor-critic algorithms and applications},
  author={Haarnoja, Tuomas and Zhou, Aurick and Hartikainen, Kristian and Tucker, George and Ha, Sehoon and Tan, Jie and Kumar, Vikash and Zhu, Henry and Gupta, Abhishek and Abbeel, Pieter and others},
  journal={arXiv preprint arXiv:1812.05905},
  year={2018}
}

@article{tirumala2023replay,
  title={Replay across experiments: A natural extension of off-policy rl},
  author={Tirumala, Dhruva and Lampe, Thomas and Chen, Jose Enrique and Haarnoja, Tuomas and Huang, Sandy and Lever, Guy and Moran, Ben and Hertweck, Tim and Hasenclever, Leonard and Riedmiller, Martin and others},
  journal={arXiv preprint arXiv:2311.15951},
  year={2023}
}

@article{justesen2025human,
  title={Human-like bots for tactical shooters using compute-efficient sensors},
  author={Justesen, Niels and Kaselimi, Maria and Snodgrass, Sam and Vozaru, Miruna and Schlegel, Matthew and others},
  journal={IEEE Transactions on Games},
  year={2025},
  publisher={IEEE}
}

@inproceedings{pearce2022counter,
  title={Counter-strike deathmatch with large-scale behavioural cloning},
  author={Pearce, Tim and Zhu, Jun},
  year=2022,
  booktitle={IEEE Conference on Games (CoG)},
}

@inproceedings{marta2025flora,
  title={FLoRA: sample-efficient preference-based RL via low-rank style adaptation of reward functions},
  author={Marta, Daniel and Holk, Simon and Vasco, Miguel and Lundell, Jens and Homberger, Timon and Busch, Finn and others},
  year=2025,
  booktitle={IEEE International Conference on Robotics and Automation (ICRA)},
}

@article{zhao2020winning,
  title={Winning is not everything: Enhancing game development with intelligent agents},
  author={Zhao, Yunqi and Borovikov, Igor and de Mesentier Silva, Fernando and Beirami, Ahmad and Rupert, Jason and Somers, Caedmon and Harder, Jesse and Kolen, John and Pinto, Jervis and Pourabolghasem, Reza and others},
  journal={IEEE Transactions on Games},
  year={2020},
}

@inproceedings{suarez2025pufferlib,
  title={Pufferlib 2.0: Reinforcement learning at 1m steps/s},
  author={Suarez, Joseph},
  booktitle={Reinforcement Learning Conference},
  year={2025}
}

@inproceedings{romeo2025speq,
  title={{SPEQ}: Offline Stabilization Phases for Efficient Q-Learning in High Update-To-Data Ratio Reinforcement Learning},
  author={Romeo, Carlo and Macaluso, Girolamo and Sestini, Alessandro and Bagdanov, Andrew D},
  booktitle={Reinforcement Learning Conference},
  year={2025}
}

@inproceedings{schwarzer2023bigger,
  title={Bigger, better, faster: Human-level atari with human-level efficiency},
  author={Schwarzer, Max and Ceron, Johan Samir Obando and Courville, Aaron and others},
  booktitle={International Conference on Machine Learning},
  year={2023},
}

@article{faldor2024omni,
  title={Omni-epic: Open-endedness via models of human notions of interestingness with environments programmed in code},
  author={Faldor, Maxence and Zhang, Jenny and Cully, Antoine and Clune, Jeff},
  journal={arXiv preprint arXiv:2405.15568},
  year={2024}
}

@article{dohare2024loss,
  title={Loss of plasticity in deep continual learning},
  author={Dohare, Shibhansh and Hernandez-Garcia, J Fernando and Lan, Qingfeng and others},
  journal={Nature},
  year={2024},
  publisher={Nature Publishing Group UK London}
}

@article{yue2026scaling,
  title={Scaling Behavior Cloning Improves Causal Reasoning: An Open Model for Real-Time Video Game Playing},
  author={Yue, Yuguang and Salia, Irakli and Hunt, Samuel and Green, Chris and Shi, Wenzhe and Hunt, Jonathan J},
  journal={arXiv preprint arXiv:2601.04575},
  year={2026}
}

@article{magne2026nitrogen,
  title={NitroGen: An Open Foundation Model for Generalist Gaming Agents},
  author={Magne, Lo{\"\i}c and Awadalla, Anas and Wang, Guanzhi and Xu, Yinzhen  and others},
  journal={arXiv preprint arXiv:2601.02427},
  year={2026}
}

@article{justesen2019deep,
  title={Deep learning for video game playing},
  author={Justesen, Niels and Bontrager, Philip and Togelius, Julian and Risi, Sebastian},
  journal={IEEE Transactions on Games},
  year={2019},
}

@article{gallotta2024large,
  title={Large language models and games: A survey and roadmap},
  author={Gallotta, Roberto and Todd, Graham and Zammit, Marvin and Earle, Sam and Liapis, Antonios and Togelius, Julian and Yannakakis, Georgios N},
  journal={IEEE Transactions on Games},
  year={2024},
  publisher={IEEE}
}

@inproceedings{ahlberg2023generating,
  title={Generating personas for games with multimodal adversarial imitation learning},
  author={Ahlberg, William and Sestini, Alessandro and Tollmar, Konrad and Gissl{\'e}n, Linus},
  booktitle={IEEE Conference on Games (CoG)},
  year={2023},
}

@inproceedings{sestini2021policy,
  title={Policy fusion for adaptive and customizable reinforcement learning agents},
  author={Sestini, Alessandro and Kuhnle, Alexander and Bagdanov, Andrew D},
  booktitle={IEEE Conference on Games (CoG)},
  year={2021},
}
\end{document}